# Policy-contingent abstraction for robust robot control

Joelle Pineau, Geoff Gordon and Sebastian Thrun
School of Computer Science
Carnegie Mellon University
Pittsburgh, PA 15213

## Abstract

This paper presents a scalable control algorithm that enables a deployed mobile robot to make high-level control decisions under full consideration of its probabilistic belief. We draw on insights from the rich literature of structured robot controllers and hierarchical MDPs to propose PolCA, a hierarchical probabilistic control algorithm which learns both subtask-specific state abstractions and policies. The resulting controller has been successfully implemented onboard a mobile robotic assistant deployed in a nursing facility. To the best of our knowledge, this work is a unique instance of applying POMDPs to high-level robotic control problems.

## 1 Introduction

Real-world robotic control problems are characterized by large numbers of states, and in some cases equally large action sets. To control a robot effectively, it is necessary to learn a good *policy*. That is, for each state of the world (or for each state of the robot's knowledge of the world), we must learn which action to take. A good policy minimizes costs over time, and is also robust both to possible stochastic action effects and to the limited observability of world state which arises from noisy and inaccurate sensors.

Our search for a robust robot controller is motivated by the Nursebot project (Montemerlo et al., 2002). This project has developed a mobile robot assistant for elderly institutionalized people. Key tasks of the robot include delivering information (reminders of appointments, medications, activities) and guiding people through their environment while interacting in socially appropriate ways. Designing a good robot controller for this domain is critical since the cost of executing the wrong command can be high. Poor action choices can cause the robot to wander off to another location in the middle of a conversation, or to tell a user to take the wrong medication. The design of the controller is complicated by the fact that much of the human-robot interaction is speech-driven. While today's recognizers yield high recognition rates for articulate speakers, elderly people often lack clear articulation or the cognitive awareness to place themselves in an appropriate position for optimal reception. Thus the controller must be robust to high noise levels when inferring, and responding to, users' requests.

Some of the most successful robot control architectures rely on structural assumptions to tackle large-scale control problems (Brooks, 1986; Arkin, 1998). The Subsumption architecture for example uses a combination of hierarchical task partitioning and task-specific state abstraction to produce scalable control systems. However it, and other similar approaches, rely on human designers to specify all structural constraints (hierarchy, abstraction) and in some cases even the policies. This can require significant time and resources, and often lead to sub-optimal solutions.

This paper describes a new probabilistic planning algorithm called PolCA (for **Pol**icy-**C**ontingent **A**bstraction). Though very much in the tradition of earlier structured robot architectures, PolCA leverages techniques from the MDP literature to formalize the framework and automatically learn task-specific abstraction and policies. PolCA uses a human-designed task hierarchy which it traverses from the bottom up, learning an abstraction function and recursively-optimal policy for each subtask along the way.

PolCA shares significant similarities with well-known hierarchical MDP algorithms (Dietterich, 2000; Andre & Russell, 2002), in terms of defining subtasks and learning policies. However PolCA improves on these in three ways which are essential for robotic problems. First, it requires less information from the human designer: he or she must specify the action hierarchy, but not the abstraction function. In our experience human experts are faster and more accurate at providing hierarchies than they are at providing state abstractions, so PolCA benefits from faster controller design and deployment. The automatic state abstraction is done using an algorithm by Dean and Givan (1997), which has not been previously used in the context of hierarchies.

Second, PolCA performs *policy-contingent* abstraction: the abstract states at higher levels of the hierarchy are left unspecified until policies at lower levels of the hierarchy are fixed. By contrast, human-designed abstraction functions are usually *policy-agnostic* (correct for all possible policies) and therefore cannot obtain as much abstraction. Humans may sometimes (accidentally or on purpose) incorporate assumptions about policies into their state abstraction functions, but because these are difficult to identify and verify, they can easily introduce bugs in the final plan.

Finally, PolCA extends easily to partially observable planning problems, which is of utmost importance for robotic problems. We call the POMDP version of the algorithm PolCA+ to avoid confusion. We present theoretical results about PolCA, consistent with earlier hierarchical MDP results, and present experiments involving both PolCA and PolCA+. In particular, we describe the performance of the PolCA+ controller onboard the robot during deployment in a nursing home near Pittsburgh, PA. While much of the techniques at the heart of PolCA+ are well-known in the MDP community, this paper shows how they can be applied to subsumption-style hierarchical robot control architectures to solve real-world robot problems.

## 2 The Nursebot Initiative

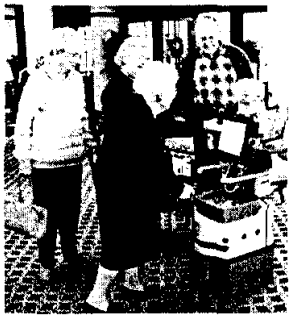

Figure 1: Pearl, the robotic nursing assistant, interacting with elderly people at a nursing facility.

The Nursebot initiative is a multi-disciplinary effort to develop a robotic nursing assistant. Pearl, the robot shown in Figure 1, has been deployed in a nursing home near Pittsburgh, PA. It employs a rich suite of navigation and recognition routines developed previously and described elsewhere (Montemerlo et al., 2002; Pollack, 2002). The robot's high-level controller, which is the focus of this paper, is concerned with decisions concerning the robot's overall behavior. Actions at this level generally involve two types of activities: (1) emitting a speech signal generated from a fixed grammar and displaying the information on the screen, and (2) moving to a designated target location. Due to the complexity of the interactive component, the number of possible actions is much larger than with a strict navigation controller. Figure 2 shows a proposed action hierarchy which reflects natural subtask groupings for this domain. [1] The state space necessary to describe the robot's task domain consists of 576 states, based on the following multi-valued features:

- robot's location (discrete approximation)
- person's location (discrete approximation)
- person's attention (inferred from speech recognizer)
- battery status (discrete approximation)
- motion goal (discrete approximation)
- reminder goal (what to inform the user of)
- user initiated goal (e.g. an information request)

Obviously, such a state space is well within the reach of present MDP algorithms. Hence, if the state was fully observable (as typically assumed in robotics), controlling the robot would be a trivial exercise. The complication arises from the fact that the state is not fully observable. The robot's sensors, most notably its speech recognition software and its location sensors, are subject to noise. For example, a robot may easily mistake phrases like "get me the time" and "get me my medicine," but whereas one involves motion, the other does not. Thus, considering uncertainty is of great importance in this domain. In particular, it is important to trade-off the costs of asking a clarification question, versus accidentally executing the wrong command. While POMDPs are perfectly equipped to address the problem of control under uncertainty, a domain of this size is well beyond the reach of existing exact solvers (Kaelbling et al., 1998). In this paper, we show how PolCA+, through its joint use of hierarchical policy constraints and automated state abstraction, is able to find a good policy for this large POMDP domain.

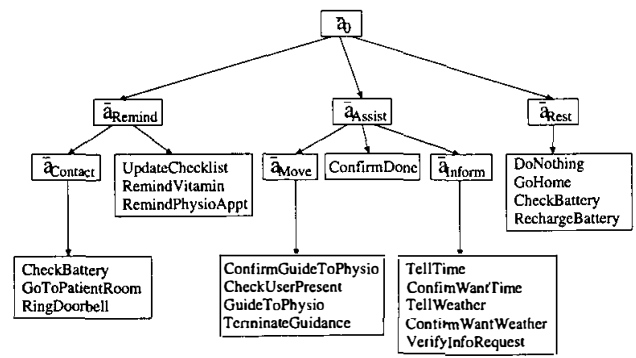

Figure 2: The Nursebot's actions.

## 3 Preliminaries

In the interest of clarity, we first describe all essential ideas in the context of PolCA (using MDPs rather than POMDPs). This allows us to draw parallels with other hierarchical MDP algorithms. Section 4.4 describes the extensions that allow PolCA+ to handle partial observability.

---
[1] Though we currently assume that the hierarchy is given, recent work has looked at automatically finding action hierarchies (McGovern & Barto, 2001; Pickett & Barto, 2002; Ryan, 2002). Most of this work does not yet apply to POMDP domains.

### 3.1 Review of MDPs

A Markov Decision Process (MDP) is a probabilistic framework to perform optimal action selection in stochastic domains. We assume the standard formulation (Kaelbling et al., 1996), where an MDP is defined to be a 4-tuple, $M = \{S, A, T, R\}$. A discrete set of states $S = \{s_0, \ldots, s_k\}$ represents the domain where a robot must select between actions $A = \{a_0, \ldots, a_n\}$. The robot aims to maximize its expected (discounted) sum of rewards: $\sum_{t=0}^{T} \gamma^t r_t(s, a)$, where $\gamma$ is the discount factor and $r_t(s, a) \in R$ is the reward received at time $t$ for executing $a$ in state $s$. The distribution $T(s', a, s)$ defines state-to-state transition probabilities, conditioned on action $a$. The main challenge of MDPs is to optimize a policy $\pi$, mapping states to actions, such that the expected sum of rewards is maximized. In small domains, the optimal policy can be found analytically by solving the Bellman equation:

$$V(s) = \max_a [R(s, a) + \gamma \sum_{s' \in S} T(s, a, s') V(s')] \quad (1)$$

$$\pi(s) = \operatorname*{argmax}_a [R(s, a) + \gamma \sum_{s' \in S} T(s, a, s') V(s')] \quad (2)$$

### 3.2 Hierarchical MDP Approaches

Hierarchical MDP problem solving accelerates planning for complex problems by leveraging domain knowledge to set intermediate goals. These define separate subtasks and constrain the solution search space. Existing hierarchical MDP approaches include MAXQ (Dietterich, 2000), HAM (Parr & Russell, 1998), ALisp (Andre & Russell, 2002), and OPTIONS (McGovern et al., 1998). Most of these approaches assume that the domain knowledge necessary to define subtasks is provided by the designer. Subtasks are formally defined using a combination of elements, including: initial states, expected goal states, fixed/partial policies, reduced action sets, and local reward functions.

In a hierarchical controller it is generally unnecessary for each subtask planner to consider every state feature. This is massively exploited in subsumption-syle robotics. By appropriately ignoring irrelevant features (or equivalently grouping similar states into a single one) we can accelerate subtask optimization without affecting policy quality. For example, in a sequence of conversation actions in the Nursebot domain, we do not need to consider the robot's precise $\{x, y\}$ location when selecting speech commands.

This idea of applying subtask-specific abstraction is well known in the hierarchical MDP literature (Dietterich, 2000; Andre & Russell, 2002). But with the exception of work by Jonsson and Barto (2001) for the OPTIONS framework, most hierarchical approaches require a hand-designed state abstraction for each subtask. As described above, hand-designed abstractions can be difficult to get right, and are unable to leverage policy-specific abstraction opportunities. For example, how coarsely can we represent the robot's $\{x, y\}$ position for conversation tasks? Clearly some abstraction is in order (e.g., all positions more than 10m from the user make conversation equally difficult), but the exact amount or form of abstraction is hard to quantify.

### 3.3 Automatic state abstraction in MDPs

There exists a related body of work that focuses on automatically discovering state abstraction functions for flat (non-hierarchical) MDPs. The overall goal there is to learn a function $z(s)$, mapping *states* to *clusters of states*, such that we can learn a policy over clusters. Since there are many ways of clustering states, the real challenge lies in finding a grouping that allows us to a) plan over clusters with minimal loss of performance, compared to planning over the entire state space, and b) significantly reduce planning time. To perform automatic abstraction, we adopt an algorithm originally proposed by Dean and Givan (1997) for (non-hierarchical) MDPs. To infer $z(s) \rightarrow C_j$, a function mapping states $s \in S$ to the set of clusters $C$:

- **Step I - Initialize clustering**: Let $z(s_i) = z(s_j)$ if

$$R(s_i, a) = R(s_j, a), \forall a \in A \quad (3)$$

- **Step II - Check stability of each cluster**: A cluster $C_i$ is deemed stable iff

$$\sum_{s' \in C_j} T(s_i, a, s') = \sum_{s' \in C_j} T(s_j, a, s'), \quad (4)$$
$$\forall (s_i, s_j) \in C_i, \forall C_j \in C, \forall a \in A$$

- **Step III - If a cluster is unstable, then split it**: Let

$$C_i \rightarrow \{C_p, \ldots, C_u\} \quad (5)$$

such that Step 2 is satisfied (and re-assign $z(s), \forall s \in C_i$).

In step I, a set of overly-general clusters are proposed. Steps II and III are then applied iteratively, gradually splitting clusters according to salient differences in model parameters, until there are no intra-cluster differences. This algorithm exhibits many desirable properties (see Dean and Givan (1997); Dean et al. (1997) for details and proofs):

1. Planning over clusters converges to optimal solution.
2. The algorithm can be relaxed to allow approximate state abstraction.
3. Given a factorized MDP, all steps can be implemented efficiently; to avoid full state space enumeration, we may occasionally miss some feasible abstractions.

While Dean and Givan's state clustering algorithm can be relaxed to allow approximate abstraction, the current version of PolCA does not include this. At first glance this may seem too strict, but it is important to realize that while exact abstraction may be very rare in flat MDPs, it is generally very prevalent in hierarchical MDPs where it stems directly from the subtask partitioning. Given that each subtask has only a small number of (related) actions, it is unsurprising that these only affect a subset of state features. For example in the Nursebot domain, the robot's position is likely unaffected by any of the actions in subtask *Inform*.

## 4 PolCA: A Hierarchical MDP Controller

PolCA builds on concepts found in both hierarchical MDP algorithms and automated state abstraction techniques. The main idea is to follow an automatic, lazy procedure, to gradually interleave planning and abstraction while traversing the hierarchy bottom-up. For each subtask, PolCA first infers appropriate parameters, then learns a state clustering function, and finally optimizes the local control policy.

The motivation behind interleaved planning and abstraction is to leverage partial policies found in lower-level tasks to automatically increase the state abstraction potential of higher-level subtasks. In the Nursebot task hierarchy (Fig. 2), once the *Inform* subtask learns to satisfy information requests, the higher-level *Assist* controller can assume that those goals will be satisfied, and thus does not need to distinguish between different information goals, or handle unsatisfied goals. The resulting increase in abstraction can have a tremendous impact on the algorithm's scalability.

Once the entire hierarchy has been traversed, the final policy is defined by the set of local subtask policies. To select actions using these local policies, we use a top-down polling approach. This means that at *every time step* we first query the policy of the top subtask; if it returns an abstract action we query the policy of that subtask, and so on down the hierarchy until a primitive action is returned. Since policy polling occurs at every time step, a subtask may be interrupted before its subgoal is reached, namely when the parent subtask suddenly selects another action.

### 4.1 Structural assumptions

PolCA relies on a set of basic structural assumptions, which are similar to other hierarchical MDP approaches. Formally, we are given a task graph $H$, where each leaf node represents a *primitive* action $a$, from the original MDP action set $A$. Each internal node has the dual role of representing both a distinct subtask (we use notation $h$ for a subtask) whose action set is defined by its immediate children in the hierarchy, as well as an *abstract* action (we use a bar, as in $\bar{a}$, to denote abstract actions) in the context of the above-level subtask. A subtask $h$ is formally defined by

- $A_h = \{\bar{a}_j, \ldots, a_p, \ldots\}$, the set of actions which are allowed in subtask $h$. Based on the hierarchy, there is one action for each immediate child of $h$.

- $\bar{R}_h(s, a)$, the local reward function. Each subtask in the hierarchy must have local (non-uniform) reward in order to optimize a local policy. In general, this is equal to the true reward $R(s, a)$. In subtasks where all available actions have equal reward (over all states), we must add a pseudo-reward to specify the desirability of satisfying the subgoal.[2]

We also require a model of the domain: $M = \{S, A, T, R\}$. Though a departure from reinforcement learning methods, this is common for automatic state abstraction. The model can be estimated from data or provided by a designer.

### 4.2 Planning with PolCA

We now provide a full description of the four steps required to learn a control policy with PolCA. Figure 3 illustrates the entire process for a simple 4-state, 2-subtask problem. The overview of the algorithm is as follows:

- Given an MDP $M = \{S, A, T, R\}$ and task hierarchy $H = \{h_0, ..., h_n\}$
- **Step 1**: Structure state space: $H \cdot S$
- For each subtask $h \in H$, in bottom-up order:
  **Step 2**: Parameterize subtask: $T(h \cdot s, a), R(h \cdot s, a)$
  **Step 3**: Cluster subtask: $h \cdot s \to z(h \cdot s)$
  **Step 4**: Solve subtask: $M_{z(h \cdot s)} \to \pi_h^*$

Step 1—*Structure state space*—uses an idea introduced in the HAM framework (Parr & Russell, 1998), that reformulates the MDP to reflect structural assumptions. The new state space $H \cdot S$ is the cross-product of the original state space $S$ and the hierarchy subtask set $H$.

Step 2—*Parameterized subtask*—appropriately translates conventional transition and reward parameters to the structured problem representation. This includes copying original transition and reward parameters from $M$ for all relevant primitive actions, as well as inferring parameters for the newly-introduced abstract actions. Given a subtask $h$, with state set $S = \{h \cdot s_0, h \cdot s_1, \ldots\}$ and action set $A_h = \{a_k, \ldots, \bar{a}_n, \ldots\}$, where $\{\bar{a}_n, \ldots\}$ invoke lower-level subtasks $\{h_n, \ldots\}$, Equations 6-9 translate the parameters from the original MDP to the structured state space.
**Case 1: Primitive actions** (e.g. state-to-state transition arrows drawn as solid line in Fig. 3), $\forall a_k \in h, a_k \in A$:

$$T(h \cdot s_i, a_k, h \cdot s_j) = T(s_i, a_k, s_j) \quad (6)$$
$$R(h \cdot s_i, a_k) = \bar{R}_h(s_i, a_k) \quad (7)$$

**Case 2: Abstract actions** (e.g. state-to-state transition arrows drawn as dotted line in Fig. 3), $\forall \bar{a}_n \in h$:

---
[2]Pseudo-rewards are unnecessary in most multi-goal robot problems, such as the Nursebot domain, where each subtask contains one or many different goals. However they are needed for some multi-step single-goal domains (Dietterich, 2000).

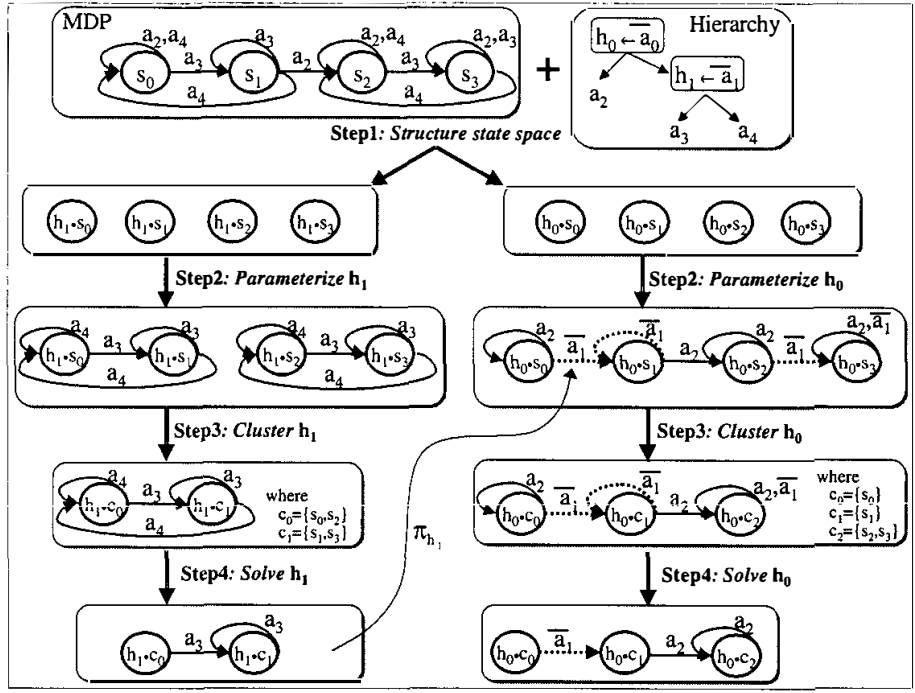

Figure 3: Simple 4-state problem with 2-subtask hierarchy. Non-zero (deterministic) transition probabilities are illustrated in the MDP. The subtask hierarchy $H$ is illustrated in the top right corner. Note that the bottom subtask ($h_1$) is solved first, and its final policy $\pi_{h_1}$ (illustrated in bottom left corner) is used to model $\bar{a}_1$ (in right column), as prescribed in Step 2.

$$T(h \cdot s_i, \bar{a}_n, h \cdot s_j) = T(s_i, \pi^*_{h_n}(s_i), s_j), \quad (8)$$

$$R(h \cdot s_i, \bar{a}_n) = \bar{R}_h(s_i, \pi^*_{h_n}(s_i)) \quad (9)$$

Equations 8-9 depend on $\pi^*_{h_n}$—the final policy of subtask $h_n$—which enforces the policy-contingent aspect of PolCA. Since parameter setting for $\bar{a}_n$ occurs *after* $h_n$ has been solved, state abstraction in $h$ need be sufficient to represent $\pi^*_{h_n}$, but not *any* policy $\pi_{h_n}$. It is worth emphasizing that because PolCA uses polling execution, abstract actions are modelled according to their *one-step* effect.[3]

Step 3—*Cluster subtask*—uses the exact iterative model minimization of Equations 3-5. Model parameters are then re-expressed over clusters:

$$T(h \cdot C_i, a, h \cdot C_j) = \sum_{s' \in C_j} T(h \cdot s, a, h \cdot s') \quad (10)$$

$$R(h \cdot C_i, a) = R(h \cdot s, a), \text{ for any } s \in C_i \quad (11)$$

Step 4—*Solve subtask*—applies dynamic programming updates over clusters:

$$V(h \cdot C_i) = \max_a [R(h \cdot C_i, a) + \gamma \sum_{C_j \in C} T(h \cdot C_i, a, h \cdot C_j) V(h \cdot C_j)] \quad (12)$$

The repeated application of this value update is guaranteed to converge. The final value function solution is contained in the value function of the top subtask: $V_{\pi^*_{h_0}}(h_0 \cdot s)$.

[3]This is in contrast to other hierarchical MDP approaches (e.g. MAXQ, ALisp, Options) which use *cumulative effects*:
$T(h \cdot s_i, \bar{a}_n, h \cdot s_j) = \sum_N P^{\pi_{h_n}}(s_j, N | s_i, \bar{a}_n)$.

### 4.3 Theoretical properties of PolCA

Most traditional subsumption-style robot control architectures offer little in terms of theoretical guarantees. However because it relies on a solid MDP framework, PolCA can offer theoretical properties similar to that of other hierarchical MDP approaches; we now discuss these.

Most hierarchical MDP approaches, as a result of imposing structural constraints, effectively reduce their policy search space, and thus cannot guarantee finding a globally optimal solution. Instead, they can guarantee weaker forms of optimality, for example *hierarchical optimality* and *recursive optimality*, as defined in (Dietterich, 2000).

The main difference between the two is a function of *context*. A recursively optimal solution guarantees optimality over a subtask's local policies, conditioned on that of its descendants. This is obtained when subtask policies are optimized without regard to the context in which each subtask is called. In contrast, hierarchical optimality guarantees optimality over the set of policies consistent with the hierarchy, which can be achieved by keeping track of all possible contexts for calling subtasks, which is key when subtasks have multiple goal states. There is a trade-off between solution quality and representation: though in some domains hierarchical optimality offers a better solution, this comes at the expense of lesser state abstraction and thus recursive optimality is generally considered more scalable.

Because it fixes low-level subtask policies prior to solving higher-level subtasks, PolCA achieves *recursive*, rather than *hierarchical* optimality. In many cases, including the Nursebot domain which does feature multi-goal subtasks, this loss is a small price to pay for the sizeable scalability benefits that come from policy-contingent abstraction.

**Theorem 1** *Let* $M = \{S, A, T, R\}$ *be an MDP and let* $H = \{h_0, \ldots, h_m\}$ *be a subtask graph with well-defined terminal states and pseudo-reward functions. Then PolCA computes* $\pi_H^*$, *a recursively optimal policy for* $M$ *that is consistent with* $H$.

**Proof:** The reader is referred to (Pineau & Thrun, 2002) for a complete proof. The main idea is that when solving each subtask, the parameters are fixed, and in the case of abstract actions, reflect true effects of the final policy for the corresponding subtask. Therefore the subtask policy must be optimal with respect to its restricted set of actions, which by definition meets the criteria for recursive optimality. This line of argument does not take into account the differences in state abstraction between related subtasks, however it is easy to show that the learned abstraction function does not have any effect on the policy optimization. ∎

### 4.4 PolCA+: Planning for Hierarchical POMDPs

POMDPs are a generalization of MDPs which allow for partial state observability (Kaelbling et al., 1998). Because of this, they are the ideal framework for modelling most robot problems. However planning in POMDPs is typically exponential in the size of the domain, making exact solutions intractable for most robot problems. Nonetheless by leveraging the structural assumptions of PolCA, and extending it to the partially observable case, we obtain a scalable hierarchical POMDP algorithm that is able to tackle real-world robot problems. We call this algorithm PolCA+. We now present the modifications necessary to handle partial observation, and thus go from PolCA to PolCA+.

First, we modify the model minimization stability criteria (Eqn. 4) to also check for similar observation probabilities. In POMDPs, a cluster $C_i$ is deemed stable iff:

$$\sum_{s' \in C_j} T(s_i, a, s') O(o, a, s') = \sum_{s' \in C_j} T(s_j, a, s') O(o, a, s'),$$

$$\forall (s_i, s_j) \in C_i, \forall C_j \in C, \forall a \in A, \forall o \in \Omega \quad (13)$$

Second, we set policy-contingent observation parameters (same as Eqns 6&8, but for $O(o, a_k, h \cdot s_j)$). Finally, we use an appropriate POMDP solver to optimize local policies. Throughout our experiments, we mostly optimize policies using the exact Incremental Pruning algorithm (Cassandra et al., 1997); for some larger domains we rely instead on the AMDP algorithm (Roy & Thrun, 2000).

Given the properties of belief space planning, PolCA+ cannot guarantee recursive optimality. Despite this, its ability to handle partial observability makes it much better suited than MDP approaches for real-world robot problems.

## 5 Experimental results

The control problem for the Nursebot domain was described in Section 2. The main challenges are the fact that because the robot's sensors are subject to substantial noise, particularly from the speech recognizer, MDP techniques are inadequate to robustly control the robot. The PolCA+ algorithm described in this paper significantly improves the tractability of POMDP planning, to the point where we can rely on POMDP-based planning for a real-world robot control domain such as the Nursebot project.

PolCA+ requires both an action hierarchy and model of the domain to proceed. The hierarchy (shown in Fig. 2) was designed by hand. Though the model could be learned from experimental data, the prohibitive cost of gathering sufficient data from our elderly users makes this an impractical solution. Therefore the model was designed by hand. It is characterized by non-deterministic transition/observation probabilities, large positive rewards for correctly delivering information or satisfying motion goals, large negative rewards for incorrect motion or reminder actions, and small negative rewards for clarification actions.

Because of the difficulties involved with conducting human subject experiments, only the final PolCA+ policy was deployed onboard the robot. Nonetheless, we can compare its performance in simulation with that of other planners. We first compare state abstraction possibilities between PolCA (which falsely assumes full observability) and PolCA+ (which considers similarity in observation probabilities before clustering states). This is a direct indicator of model reduction potential, and equivalently planning time. Figure 4a shows significant model compression for both PolCA or PolCA+, compared to the no-abstraction case (*NoAbs*). Differences between PolCA and PolCA+ arise when certain state features, though independent with respect to transitions and rewards, become correlated during belief tracking through the observation probabilities.

Second, we compare the reward gathered over time by each policy. As shown in Figure 4b, PolCA+ clearly outperforms PolCA in this respect. A closer look at the performance of PolCA reveals that it often answers a wrong query because it is unable to appropriately select amongst clarification actions. In other instances, the robot prematurely terminates an interaction before the goal is met, because the controller is unable to ask the user whether s/he is done. In contrast, PolCA+ resorts to confirmation actions to avoid wrong actions, and satisfy more goals. Also included in this comparison is QMDP, a fast approximate POMDP algorithm (Littman et al., 1995). On this task, it performs particularly poorly, repeatedly selecting to *doNothing* because of its inability to selectively gather information.

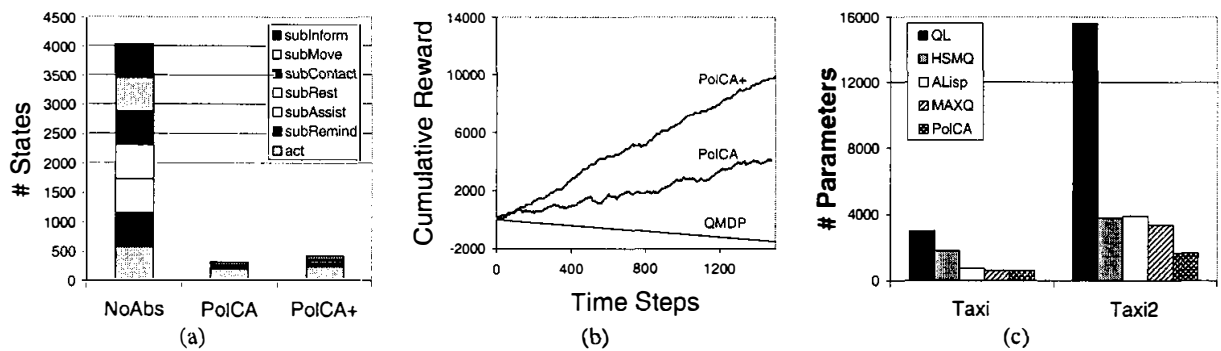

Figure 4: (a) Number of parameters for Nursebot domain. (b) Cumulative reward over time in Nursebot domain. (c) Number of Q-values to learn solution in Taxi domain. Q-learning (QL) and HSMQ (Dietterich, 2000) parameterization are included as a base comparison. These results assume a clustering function on (state,action) pairs, instead of on states. This is standard for MDP compression, including MAXQ, ALisp and PolCA (see (Pineau & Thrun, 2002) for details). However it does not carry-over to POMDPs (or PolCA+).

In terms of computation time, PolCA+ reached its solution in 18 minutes. In comparison, Incremental Pruning (an exact POMDP solver), could only complete 2 iterations of VI in 24hrs, and thus would probably take many years to reach a reasonable solution. The many-fold improvement found in PolCA+ is for the most part due to its structural assumptions (hierarchy+abstraction); some of the improvement is also achieved by using an AMDP solver at the highest-level of the hierarchy (lower subtasks are solved exactly). The PolCA solution was computed in only 4 seconds, whereas the (unstructured) QMDP solution took 39 seconds.

Table 1: A sample dialogue with a test subject. Actions in bold font are clarification actions, generated by the POMDP because of high uncertainty.

| Observation | Action | Reward |
|---|---|---|
| (null) | DoNothing | -1 |
| Scheduler=*physio-message* | gotoPatientRoom | 5 |
| Laser=*RobotAtPatientRoom* | RingBell | 5 |
| Speech=*yes* | RemindPhysioAppt | 50 |
| Speech=*noise* | **ConfirmGuideToPhysio** | -5 |
| Speech=*yes* | **CheckBattery** | -5 |
| Battery=*high* | GuideToPhysio | 50 |
| Laser=*RobotAtPhysio* | CheckUserPresent | -1 |
| IR=*no-user* | CheckUserPresent | -1 |
| IR=*user* | CheckUserPresent | -5 |
| IR=*user* | TerminateGuidance | 50 |
| Speech=*noise* | ConfirmDone | -1 |
| Speech=*no* | **VerifyInfoRequest** | -1 |
| Speech=*weather* | ConfirmWantWeather | -1 |
| Speech=*noise* | **VerifyInfoRequest** | -1 |
| Speech=*weather* | ConfirmWantWeather | -1 |
| Speech=*yes* | TellWeather | 50 |
| Speech=*noise* | ConfirmDone | -1 |
| Speech=*yes* | GoHome | 5 |
| Laser=*RobotAtHome* | RechargeBattery | 20 |

The PolCA+ policy was implemented onboard the robot and tested during two days of experiments with elderly residents at a local nursing home. Through the course of twelve interaction scenarios, Pearl was able to successfully deliver scheduled reminders, guide residents to physiotherapy appointments, and satisfy information requests. The robustness of the PolCA+ policy to uncertainty was demonstrated by its use of clarification questions whenever a user's intentions were unclear. As a result, all six test subjects were able to complete the full experimental scenario, after receiving only limited training (a five minute introduction session). Table 5 shows a typical interaction between the robot and user, in terms of the observations received by the controller, and the actions selected in response, as well as the corresponding reward signals. Step-by-step images corresponding to this interaction are shown in Figure 5.

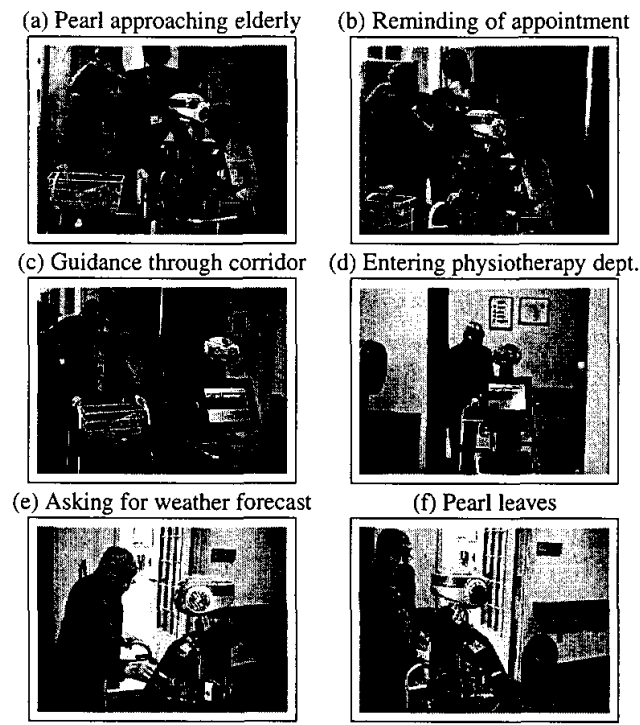

(a) Pearl approaching elderly  (b) Reminding of appointment
(c) Guidance through corridor  (d) Entering physiotherapy dept.
(e) Asking for weather forecast  (f) Pearl leaves

Figure 5: Example of a successful guidance experiment.

## 6 Comparison with hierarchical MDPs

We conclude this paper by presenting a comparison of PolCA with two competing hierarchical MDP algorithms: MAXQ and ALisp. For this, we select the Taxi domain, a commonly used problem in the hierarchical MDP literature (Dietterich, 2000). Both MAXQ and ALisp have pub-lished results for this task. The overall task (see Fig. 6a) is to control a taxi agent with the goal of picking up a passenger from a start location, and then dropping him/her off at their destination. We include a second larger domain (Taxi2). It is identical to the first task except that the passenger can start from any location on the grid, compared to only {Y,B,R,G} in the original Taxi domain. Figure 6b represents the action hierarchy used by all algorithms for both domains.

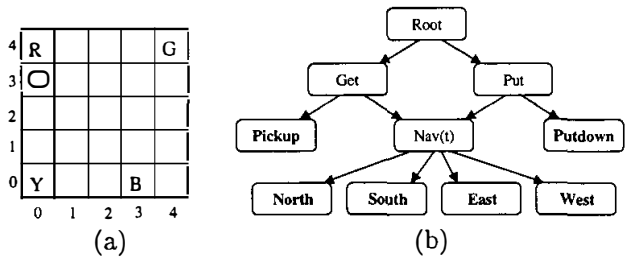

(a) (b)

Figure 6: The Taxi domain is represented using four features: {X,Y,Passenger,Destination}. The X,Y represent the taxi's position in the 5x5 grid; the passenger can be at any of: {Y,B,R,G,taxi}; the destination is one of: {Y,B,R,G}. The taxi agent can select from six actions: {N,S,E,W,Pickup,Putdown}.

Figure 4c compares state clustering results for these two tasks. In both cases, the clustering function for PolCA was learned automatically, whereas that for MAXQ and AL-isp was hand-crafted. These results confirm the fact that performing automatic abstraction can allow further model reduction; most of the gains can be attributed to the policy-contingent aspect of PolCA. This occurs despite the fact that PolCA does not use a decomposed value function. All algorithms learn the optimal policy for both tasks.

## 7 Conclusion

This paper describes PolCA+, a novel algorithm for scalable POMDP planning. By combining techniques from hierarchical MDPs and automated state abstraction, and including a straight-forward extension to POMDPs, we are able to produce a planning algorithm capable of performing high-level control of a mobile interactive robot. It is the first hierarchical POMDP algorithm to do so, and was a key element for the successful performance of the robot in a series of experiments with elderly users. While we agree with the long-held view in the robot community that structural assumptions are a necessary ingredient to address large-scale problems, we believe PolCA+'s ability to perform automated state abstraction and policy learning, as well as handle uncertainty, are significant improvements over earlier robot architectures.